%% file: main.tex
\documentclass[11pt,a4paper]{article}
\usepackage[hyperref]{acl2021}
\usepackage{times}
\usepackage{latexsym}

\usepackage{microtype}

\usepackage[ruled,linesnumbered]{algorithm2e}
\usepackage{subfigure}
\usepackage{enumitem}
\usepackage{multirow}
\usepackage{booktabs}
\usepackage{color}
\usepackage{amsmath}
\usepackage{algorithmic}
\usepackage{ulem}
\usepackage{graphicx}
\usepackage{amssymb}

\aclfinalcopy 

\title{The Authors Matter: \\ Understanding and Mitigating Implicit Bias in Deep Text Classification}

\author{
	{Haochen Liu\textsuperscript{1}, Wei Jin\textsuperscript{1}, Hamid Karimi\textsuperscript{1}, Zitao Liu\textsuperscript{2}, Jiliang Tang\textsuperscript{1}}
	\vspace{1.6mm}\\
	\fontsize{12}{10}\selectfont\itshape
	\,\textsuperscript{\rm 1}  Michigan State University, East Lansing, MI, USA \\
	\fontsize{12}{10}\selectfont\itshape  \textsuperscript{\rm 2} TAL Education Group, Beijing, China \\
    \fontsize{10}{10}\selectfont \{liuhaoc1,jinwei2,karimiha\}@msu.edu;  liuzitao@100tal.com; tangjili@msu.edu\\}

\date{}

\begin{document}
\maketitle
\begin{abstract}
\input{abs}
\end{abstract}

\input{intro}
\input{prelim}

\input{bias_interpretation}

\input{bias_mitigation}
\input{experiments}

\input{relatedwork}
\input{conclusion}



\bibliographystyle{acl_natbib}
\bibliography{main}

\appendix
\input{appendix}

\end{document}

%% file: abs.tex
It is evident that deep text classification models trained on human data could be biased. In particular, they produce biased outcomes for texts that explicitly include identity terms of certain demographic groups. We refer to this type of bias as explicit bias, which has been extensively studied. However, deep text classification models can also produce biased outcomes for texts written by authors of certain demographic groups. We refer to such bias as implicit bias of which we still have a rather limited understanding. In this paper, we first demonstrate that implicit bias exists in different text classification tasks for different demographic groups. Then, we build a learning-based interpretation method to deepen our knowledge of implicit bias. Specifically, we verify that classifiers learn to make predictions based on language features that are related to the demographic attributes of the authors. Next, we propose a framework \textbf{Debiased-TC} to train deep text classifiers to make predictions on the right features and consequently mitigate implicit bias. We conduct extensive experiments on three real-world datasets. The results show that the text classification models trained under our proposed framework outperform traditional models significantly in terms of fairness, and also slightly in terms of classification performance. 


%% file: intro.tex
\section{Introduction}

Many recent studies have suggested that machine learning algorithms can learn social prejudices from data produced by humans, and thereby show systemic bias in performance towards specific demographic groups or individuals \cite{mehrabi2019survey,blodgett2020language,DBLP:conf/acl/ShahSH20}.
As one machine learning application, text classification has been proven to be discriminatory towards certain groups of people \cite{dixon2018measuring,borkan2019nuanced}. Text classification applications such as sentiment analysis and hate speech detection are common and widely used in our daily lives. If a biased hate speech detection model is deployed by a social media service provider to filter users' comments, the comments related to different demographic groups can have uneven chances to be recognized and removed. Such a case will cause unfairness and bring in negative experience to its users. Thus, it is highly desired to mitigate the bias in text classification.

\begin{table*}[t]
    \small
	\centering
	\caption{An illustrative example on the implicit bias of a CNN text classification model.}
	\vspace{-0.1in}
	\begin{tabular}{c|l||c|c}
	    \hline \hline
		Author & Text & Label & Prediction \\ \hline
		
		White American & \colorbox{red!20}{Can't} \colorbox{red!30}{wait} to visit \colorbox{red!50}{your} new home. Yes, I going to be \colorbox{red!40}{a} \colorbox{red!70}{great} guest! & positive & positive \\
\hline
		
		African American & \colorbox{red!30}{Can't} \colorbox{red!30}{wait} to visit \colorbox{red!50}{your} new home. \colorbox{red!35}{Yup}, I \colorbox{red!20}{goin} to be a \colorbox{red!40}{great} guest! & positive & negative \\
		\hline \hline

	\end{tabular}
	\vspace{-0.2in}
	\label{tab:example}
\end{table*}

The majority of existing studies on bias and fairness in text classification have mainly focused on the bias towards the individuals mentioned in the text content. For example, in~\cite{dixon2018measuring,park2018reducing,zhang2020demographics}, it is investigated how text classification models perform unfairly on texts containing demographic identity terms such as ``gay'' and ``muslim''. In such scenarios, the demographic attributes of the individuals subject to bias explicitly exist in the text. In this work, we refer to this kind of bias as \textbf{explicit bias}. Bias in texts, however, can be reflected more subtly and insidiously. While a text may not contain any reference to a specific group or individual, the content can somehow be revealing of the demographic information of the author. As shown in \cite{coulmas2013sociolinguistics,preoctiuc2018user}, the language style (e.g., wordings and tone) of a text can be highly correlated with its author's demographic attributes (e.g., age, gender, and race).
We find that a text classifier can learn to associate the content with demographic information and consequently make unfair decisions towards certain groups.
We refer to such bias as \textbf{implicit bias}. Table \ref{tab:example} demonstrates an example of implicit bias. There are two short texts where the first text is written by a white American and the second one by an African American. The task is to predict the sentiment of a text by a convolutional neural network (CNN) model. Words with a red background indicate those with the salient predictive capability by the model where the darker the color, the more salient the words. The words ``yup'' and ``goin'' in the second text are commonly used by African Americans \cite{liu2020does} and are irrelevant to the sentiment. However, the CNN model has hinted at them and consequently has predicted a positive text to be negative.

In this work, we aim to understand and mitigate implicit bias in deep text classification models. One key source of bias is the imbalance of training data~\cite{dixon2018measuring,park2018reducing}. Thus, existing debiasing methods mainly focus on balancing the training data, such as adding new training data \cite{dixon2018measuring} and augmenting data based on identity-term swap \cite{park2018reducing}. However, these methods cannot be directly applied to mitigate implicit bias. Obtaining new texts from authors of various demographic groups is very expensive. It requires heavy human labor. Meanwhile, given that there is no explicit demographic information in texts, identity-term swap data augmentation is not applicable. Thus, we propose to enhance deep text classification models to mitigate implicit bias in the training process. To achieve this goal, we face tremendous challenges. First, to mitigate the implicit bias, we have to understand how deep models behave. For example, how they correlate implicit features in text with demographic attributes and how the models make biased predictions. Second, we need to design new mechanisms to take advantage of our understandings to mitigate the implicit bias in deep text classifiers. 
 						
To address the above challenges, in this paper, we first propose an interpretation method, which sheds light on the formation mechanism of implicit bias in deep text classification models. We show that the implicit bias is caused by the fact that the models make predictions based on incorrect language features in texts. Second, based on this finding, we propose a novel framework \textbf{Debiased-TC} (Debiased Text Classification) to mitigate the implicit bias of deep text classifiers. More specifically, we equip the deep classifiers with an additional saliency selection layer that first determines the correct language features which the model should base on to make predictions. We also propose an optimization method to train the classifiers with the saliency selection layer. Note that both our proposed interpretation method and the learning framework are model-agnostic, where they can be applied to any deep text classifier. We evaluate the framework with two popular deep text classification models across various text classification tasks on three public datasets. The experimental results demonstrate that our method significantly mitigates the implicit bias while maintaining or even improving their prediction performance.



%% file: prelim.tex
\section{Preliminary Study}
\label{sec:prelim}
In this section, we perform a preliminary study to validate the existence of implicit bias in deep text classification models. We first introduce the data and text classification tasks, and then present the empirical results. 

\subsection{Data and Tasks}
In this preliminary study, we investigate different text classification tasks and various demographic groups to validate the implicit bias. We use three datasets, including the DIAL and PAN16 datasets processed by \cite{elazar2018adversarial} and the Multilingual Twitter Corpus (MTC) introduced in \cite{huang2020multilingual}.


The DIAL dataset contains dialectal texts collected from Twitter. Each tweet's text is associated with the \textit{race} of the author as the demographic attribute, denoted as ``white'' and ``black'', respectively. This dataset is annotated for two classification tasks: sentiment analysis and mention detection. The sentiment analysis task aims to categorize a text as ``happy'' or ``sad''. The mention detection task tries to determine whether a tweet mentions another user, which can also be viewed as distinguishing conversational tweets from non-conversational ones.

The PAN16 dataset consists of tweets. For each tweet, \textit{age} and \textit{gender} of its author have been manually labelled. The demographic attribute age has two categories of ($18$-$34$) and ($\geq 35$), and gender has male and female. Also, this dataset is annotated for the mention detection task as described above.

The MTC dataset contains multilingual tweets for the hate speech detection task. Each tweet is annotated as ``hate speech'' or ``non hate speech'' and associated with four author's demographic attributes: race, gender, age, and country. We only use the English corpus with the attribute \textit{race}. In this dataset, the attribute race has two categories, i.e., white and nonwhite.

More statistical information on these three datasets and the links to downloadable versions of the data can be found in Appendix \ref{app:sta}.


\subsection{Empirical study}
\label{sec:prelim_results}

In this subsection, we aim to empirically study if text classification models make the predictions dependent on the demographic attributes of the authors of the texts. The explicit bias in text classification tasks stems from the imbalance of training data~\cite{dixon2018measuring,park2018reducing}. For example, when there are more negative examples from one group in the training data, the model learns to correlate that group with the negative label, which results in bias. Inspired by this observation, to validate the existence of implicit bias, we investigate if the imbalance of training data in terms of demographic attributes of the authors can lead to biased predictions. To answer this question, we consider the following setting: (1) the training data has an equal number of positive and negative examples; and (2) positive and negative examples in the training data are imbalanced among different groups of the authors according to their demographic attributes. Intuitively, if the predictions are independent of the demographic attributes of authors, the model should still perform similarly for different groups.


For each task and demographic attribute of authors, we consider two labels (i.e., positive and negative) and two demographic groups (i.e., Group I and Group II).
For each dataset, we follow the aforementioned setting to build a training set. We make the training set overall balanced in terms of the labels and demographic groups. That is, we set the overall ratio of positive and negative examples as 1:1, and the overall ratio of examples from Group I and Group II as 1:1 as well. Meanwhile, we make the data in each group imbalanced. In particular, for Group I, we set the ratio of its positive and negative examples to 4:1, while the ratio is automatically set to 1:4 for Group II. We name the proportion of positive and negative samples in Group I as ``balance rate''.
We train a convolution neural network (CNN) text classifier as a representative model on the training set and evaluate it on the test set. 
We use the false positive/negative rates \cite{dixon2018measuring} and the demographic parity rate (a.k.a., positive outcome rate, the probability of the model predicting a positive outcome for one group) \cite{dwork2012fairness,kusner2017counterfactual} to evaluate the fairness of the classification models.



\begin{table*}[t]
\small
\centering
\caption{Preliminary study.}
\vspace{-0.1in}
\label{tab:prelim}
\begin{tabular}{ccccccccc}
\hline
\hline
\multirow{2}{1cm}{\textbf{Dataset}} & \multirow{2}{1cm}{\textbf{Task}} & \multirow{2}{1cm}{\textbf{Demo}} & \multicolumn{2}{c}{\textbf{\begin{tabular}[c]{@{}c@{}}False Positive (\%) \end{tabular}}} & \multicolumn{2}{c}{\textbf{\begin{tabular}[c]{@{}c@{}}False Negative (\%) \end{tabular}}} & \multicolumn{2}{c}{\textbf{\begin{tabular}[c]{@{}c@{}}Demographic Parity (\%) \end{tabular}}} \\ \cmidrule(r){4-5}  \cmidrule(r){6-7} \cmidrule(r){8-9}
 & & & \textbf{Group I} & \textbf{Group II} & \textbf{Group I} & \textbf{Group II} & \textbf{Group I} & \textbf{Group II} \\ \hline
\multirow{2}{1cm}{\textbf{DIAL}} & \textbf{Sentiment} & \textbf{Race} & 46.97 & 23.38 & 21.29 & 62.75 & 62.84 & 30.32 \\
 & \textbf{Mention} & \textbf{Race} & 48.72 & 15.99 & 17.32 & 34.90 & 65.70 & 40.55 \\
 \hline
\multirow{2}{1cm}{\textbf{PAN16}} & \textbf{Mention} & \textbf{Gender} & 23.90 & 12.30 & 13.06 & 23.01 & 55.42 & 44.64 \\
 & \textbf{Mention} & \textbf{Age} & 24.91 & 9.88 & 16.48 & 26.43 & 54.22 & 41.72 \\
 \hline
\textbf{MTC} & \textbf{Hate Speech} & \textbf{Race} & 80.33 & 1.77 & 12.13 & 49.35 & 84.10 & 26.21 \\
\hline
\hline
\vspace{-0.3in}
\end{tabular}
\end{table*}

The results are shown in Table \ref{tab:prelim}. For the demographic attribute race, Group I/Group II stands for white/black in the DIAL dataset, and white/nonwhite in the MTC dataset. For gender and age, Group I/Group II stands for male/female and age ranges (18-34)/($\geq$35), respectively.  From the table, we observe that in terms of different tasks and demographic attributes of authors, the model shows significant bias with the same pattern. For all cases, the demographic group with more positive examples (Group I) always gets a higher false positive rate, a lower false negative rate, and a higher demographic parity rate than the other group. This demonstrates that imbalanced data can cause implicit bias, and the predictions are not independent of the demographic attributes of authors. Since the text itself doesn't explicitly contain any demographic information, the model could learn to recognize the demographic attributes of authors based on implicit features such as language styles and associate them with a biased outcome. Next, we will understand one formation of implicit bias and then propose Debiased-TC to mitigate it.



%% file: bias_interpretation.tex
\section{Understanding Implicit Bias}
\label{sec:bias_interpretation}
In this section, we aim to understand the possible underlying formation mechanism of the implicit bias. Our intuition is -- when a training set for sentiment analysis has more positive examples from white authors and more negative examples from black authors, a classification model trained on such a dataset may learn a ``shortcut''~\cite{mahabadi2020end} to indiscriminately associates the language style features of white people with the positive sentiment and those of black people with the negative sentiment. In other words, the model does not use the correct language features (e.g., emotional words) to make the prediction. Thus, we attempt to examine the following hypothesis: \textit{A deep text classification model presents implicit bias since it makes predictions based on language features that should be irrelevant to the classification task but are correlated with a certain demographic group of authors.} To verify this hypothesis, we first propose an interpretation method to detect the salient words a text classification model relies on to make the prediction. The interpretation model enables us to check the overlapping between the salient words and the words related to the authors' demographic attributes. Consequently, it allows us to understand the relationship between such overlapping and the model's implicit bias.

\subsection{An Interpretation Method}

We follow the idea of the learning-based interpretation method L2X \cite{DBLP:conf/icml/ChenSWJ18} to train an explainer to interpret a given model. The reasons for choosing L2X are -- 1) as a learning-based explainer, it learns to globally explain the behavior of a model, instead of explaining a single instance at one time; and 2) the explainer has the potential to be integrated into our debiasing framework to mitigate implicit bias in an end-to-end manner, which will be introduced in Section \ref{sec:bias_mitigation}.


A binary text classification model $\mathcal{M}: X \rightarrow Y$ maps an input text $X = (x_1, x_2, \dots, x_n)$ to a label $Y \in \{0,1\}$. For a certain model $\mathcal{M}$, we seek to specify the contribution of each word in $X$ for $\mathcal{M}$ to make the prediction $Y$. The contributions can be denoted as a saliency distribution $S= (s_1, s_2, \dots, s_n)$, where $s_i$ is the saliency score of the word $x_i$, and $\sum_{i=1}^n s_i = 1$. Given a model $\mathcal{M}$, we train an explainer $\mathcal{E}^{\mathcal{M}}: X \rightarrow S$ to estimate the saliency distribution $S$ of an input text $X$.

The explainer is trained by maximizing $I(X_S,Y)$, the mutual information \cite{cover1999elements} between the response variable $Y$ and the selected feature $X_S$ of $X$ under saliency distribution $S$. The selected feature $X_S=X \odot S = (s_1 \cdot x_1, s_2 \cdot x_2, \dots, s_n \cdot x_n)$ \footnote{Without confusion, we use $x_i$ to denote both a word and its word embedding vector.} is calculated as the element-wise product between $X$ and $S$. In our implementation, we parametrize the explainer by a bi-directional RNN followed by a linear layer and a Softmax layer. More details about the optimization of the explainer can be found in Appendix \ref{app:exp}.

\subsection{Saliency Correlation Measurement}

In this work, we assume that the text classification task is totally independent of the demographic attribute of the author of the text. In other words, language features that reflect the author's demographic information should not be taken as evidence for the main task. Thus, we propose to understand the implicit bias of a deep text classification model by examining the overlapping between salient words for the main task and the words correlated with the demographic attribute.


\begin{figure}[t]
\begin{center}
\includegraphics[scale=0.4]{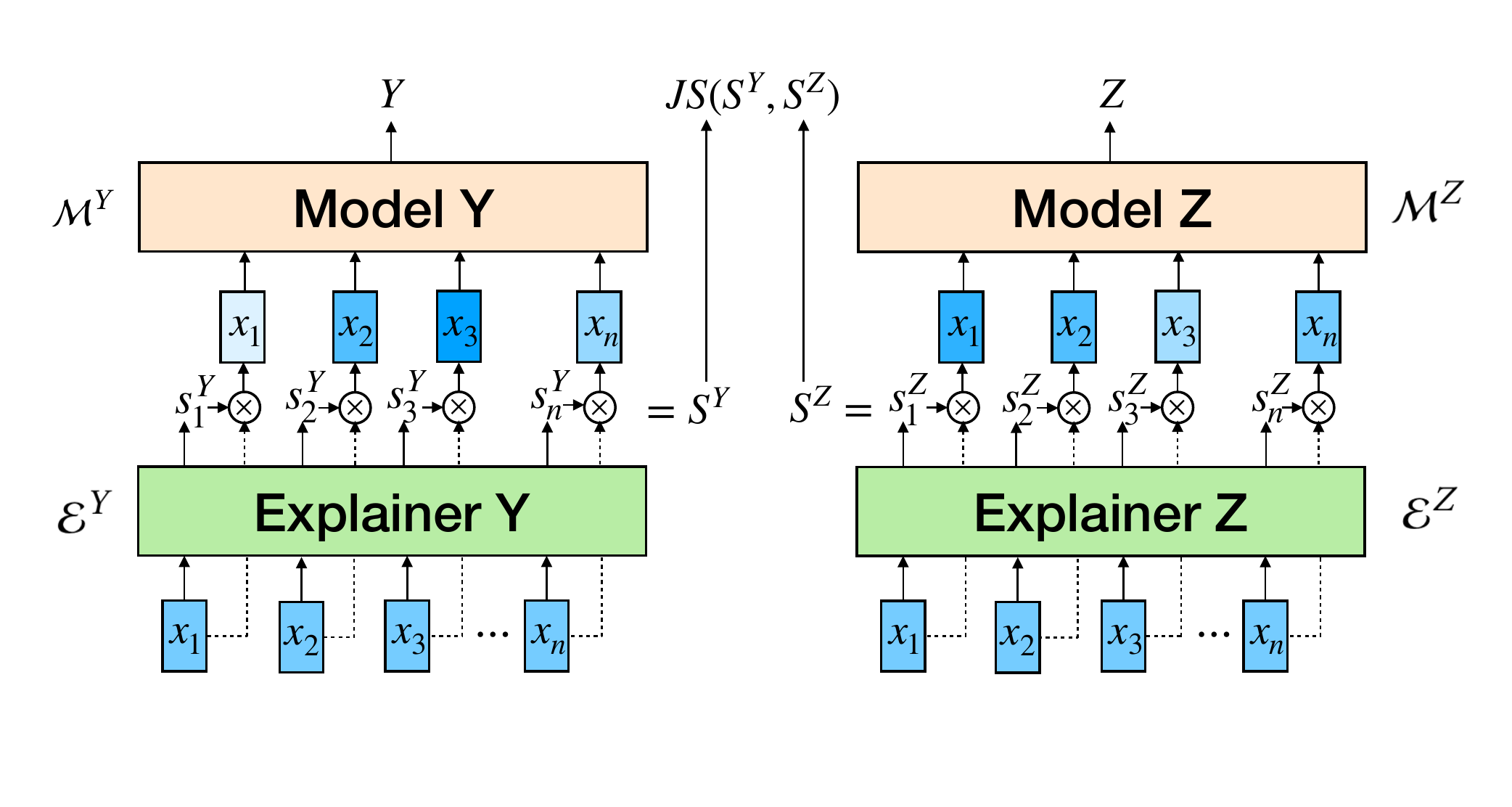}
\end{center}
\vspace{-0.8cm}
\caption{An illustration of the bias interpretation model.} 
\vspace{-0.3in}
\label{fig:interpretation}
\end{figure}

With the interpretation model, we can estimate the saliency distributions of the input words for the classification task and the demographic attribute prediction task, respectively, and then check their overlapping. As shown in Figure \ref{fig:interpretation}, we train two models $\mathcal{M}^Y$ and $\mathcal{M}^Z$ with the same architecture for the former and the latter tasks, respectively. Then, two corresponding explainers $\mathcal{E}^Y$ and $\mathcal{E}^Z$ are trained for them. Thus, given an input text $X$, two explainers can estimate the saliency distributions $S^Y$ and $S^Z$ on two tasks, respectively. We use the Jensen-Shannon (JS) divergence $JS(S^Y||S^Z)$ to measure the overlap between language features that these two tasks relying on to make the predictions on $Y$ and $Z$.

\subsection{Empirical Analysis}
\label{sec:interpret_results}

\begin{figure}[t]
\begin{center}
\includegraphics[scale=0.9]{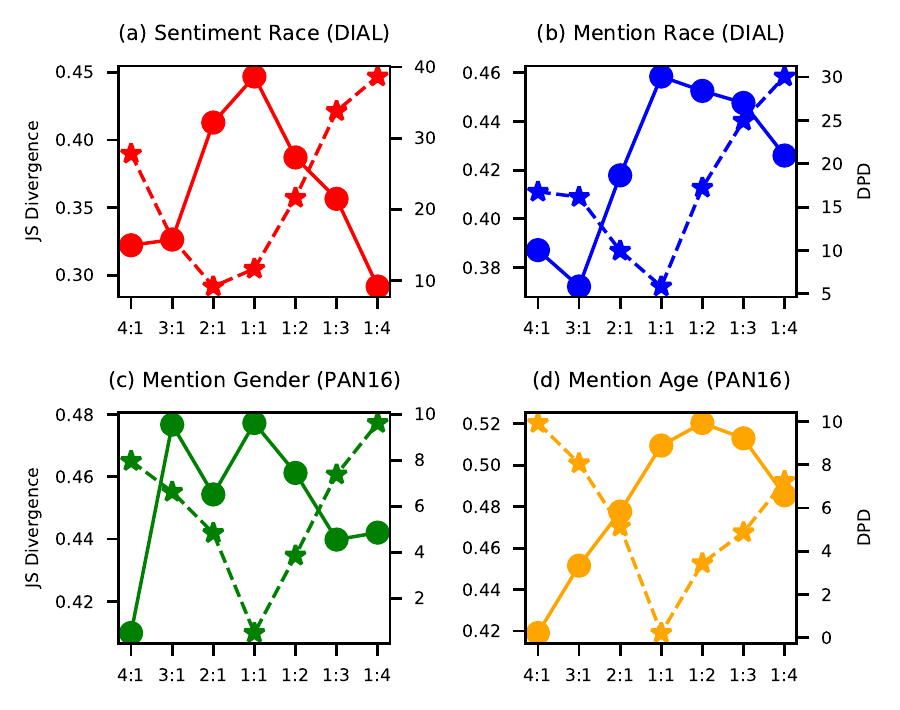}
\end{center}
\vspace{-0.4cm}
\caption{The average JS divergence (solid lines) and DPD (dash lines) vs. the balance rate. The x-axis indicates the balance rate of the training set. The y-axis on the left hand indicates the average JS divergence, and the y-axis on the right hand is the DPD.}
\vspace{-0.2in}
\label{fig:JS}
\end{figure}

In this subsection, we present the experiments to verify our hypothesis on the formulation of implicit bias. Following the experimental settings in Section \ref{sec:prelim_results}, we vary the ``balance rate'' of the training data and then observe how the saliency correlation changes. We use CNN text classifiers (see \ref{sec:settings} for details) for both $\mathcal{M}^Y$ and $\mathcal{M}^Z$. In Figure \ref{fig:JS}, we show how the average JS divergence and the demographic parity difference (DPD) vary with the changes of the balance rate. DPD is the absolute value of the difference between the demographic parity rates for the two groups. We only report the results for DIAL and PAN16 datasets and DPD as the fairness metric since we achieved similar results for other settings. For each task and each demographic attribute, the DPD is small when the training data are balanced and becomes large when the data are imbalanced. However, the JS divergence is large for balanced data while small for imbalanced data. A larger DPD indicates stronger implicit bias and a smaller JS divergence stands for a stronger overlap between the saliency distributions for the two tasks. Thus, these observations suggest that when the training data are imbalanced, the text classifiers tend to use language features related to the demographic attribute of authors to make the prediction.

%% file: bias_mitigation.tex
\section{The Bias Mitigation Framework}
\label{sec:bias_mitigation}

In the previous section, we showed that a model with implicit bias tends to utilize features related to the demographic attribute of authors to make the prediction, especially when training data is imbalanced in terms of the demographic attribute of authors. One potential solution is to balance the training data by augmenting more examples from underrepresented groups. However, collecting new data from authors of different demographics is expensive. Thus, to mitigate the implicit bias, we propose the novel framework \textbf{Debiased-TC}. Our proposed approach can mitigate implicit bias by automatically correcting their selection of input features. In this section, we will first introduce the proposed framework with the corresponding optimization method.

\begin{figure}[t]
\begin{center}
\includegraphics[scale=0.45]{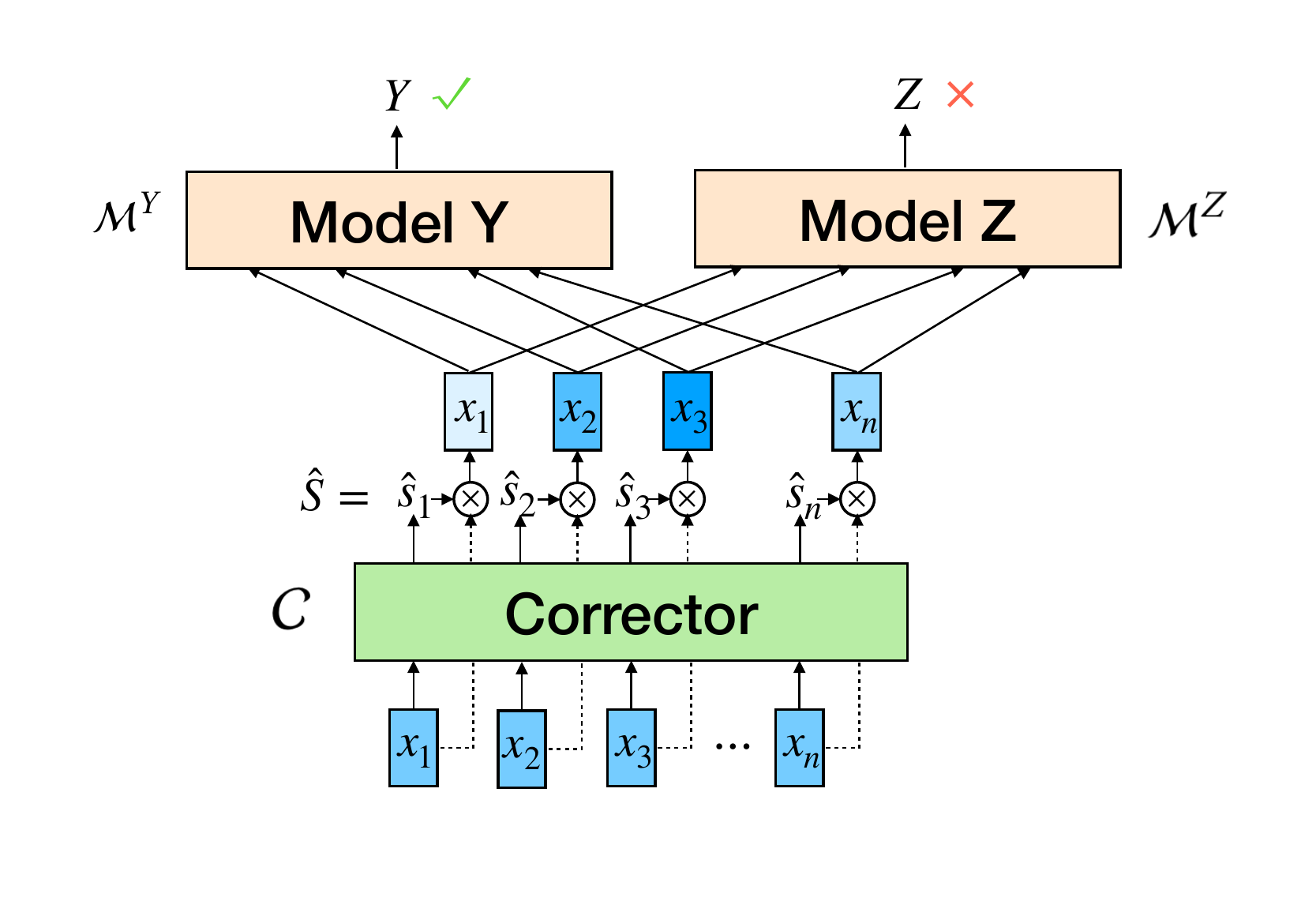}
\end{center}
\vspace{-0.8cm}
\caption{An illustration of the bias mitigation model.} 
\vspace{-0.1in}
\label{fig:debias}
\end{figure}

\subsection{Debiased Text Classification Model}
An illustration of \textbf{Debiased-TC} is shown in Figure \ref{fig:debias}. Similar to the explainer in the interpretation model, we equip the base model $\mathcal{M}^Y$ with a corrector layer $\mathcal{C}$ after the input layer. The corrector $\mathcal{C}: X \rightarrow S$ learns to correct the model's feature selection. It first maps an input text $X=(x_1, x_2, \dots, x_n)$ to a saliency distribution $S=(s_1, s_2, \dots, s_n)$, which is expected to give high scores to words related to the main tasks and low scores to words related to demographic attributes of authors. Then, it assigns weights to the input features with the saliency scores by calculating $X_{S} = X \odot S$, which is fed into the classification model $\mathcal{M}^Y$ for prediction.

%

To train a corrector to achieve the expected goal, we adopt the idea of adversarial training. More specifically, in addition to the main classifier $\mathcal{M}^Y$, we introduce an adversarial classifier $\mathcal{M}^Z$, which takes $X_{S}$ as the input and predicts the demographic attribute $Z$. During the adversarial training, the corrector attempts to help  $\mathcal{M}^Y$ make correct predictions while preventing $\mathcal{M}^Z$ from predicting demographic attributes. To make this feasible, we use the gradient reversal technique \cite{ganin2015unsupervised}, where we add a gradient-reversal layer between the weighted inputs $X_{S}$ and the adversarial classifier $\mathcal{M}^Z$. The gradient-reversal layer has no effect on its downstream components (i.e., the adversarial classifier $\mathcal{M}^Z$). However, during back-propagation, the gradients that pass down through this layer to its upstream components (i.e., the corrector $\mathcal{C}$) are getting reversed. As a result, the corrector $\mathcal{C}$ receives opposite gradients from $\mathcal{M}^Z$. The outputs of the $\mathcal{M}^Y$ and $\mathcal{M}^Z$ are used as signals to train the corrector such that it can upweight the words correlated with the main task label $Y$ and downweight the words correlated with the demographic attribute $Z$. We set the adversarial classifier $\mathcal{M}^Z$ with the same architecture as the main classifier $\mathcal{M}^Y$. The corrector $\mathcal{C}$ has the same architecture as the explainer introduced in Section \ref{sec:bias_interpretation}.

\subsection{An Optimization Method for Debiased-TC}

In this subsection, we discuss the optimization method for the proposed framework. We denote the parameters of $\mathcal{M}^Y$, $\mathcal{M}^Z$ and $\mathcal{C}$ as $\mathbf{W}^Y$, $\mathbf{W}^Z$ and $\mathbf{\Theta}$, respectively. The optimization task is to jointly optimize the parameters of the classifiers, i.e., $\mathbf{W}^Y$ and $\mathbf{W}^Z$, and the parameters of the corrector, i.e., $\mathbf{\Theta}$. We can view the optimization as an architecture search problem. Since our debiasing framework is end-to-end and differentiable, we develop an optimization method for our framework based on the differentiable architecture search (DARTS) techniques~\cite{liu2018darts,zhao2020memory}. We update $\mathcal{M}^Y$, $\mathcal{M}^Z$ by optimizing the training losses $L_{train}^Y$ and $L_{train}^Z$ on the training set and update $\mathbf{\Theta}$ by optimizing the validation loss $L_{val}$ on the validation set through gradient descent. We denote the cross-entropy losses for $\mathcal{M}^Y$ and $\mathcal{M}^Z$ as $L^Y$ and $L^Z$, respectively. $L_{train}^Y$ and $L_{train}^Z$ indicate the cross-entropy losses $L^Y$ and $L^Z$ on the training set. $L_{val}$ denotes the combined loss of the two cross-entropy losses $L=L^Y + L^Z$ on the validation set. 

The goal of optimizing the corrector is to find optimal parameters $\mathbf{\Theta}^*$ that minimizes the validation loss $L_{val} (\mathbf{W}^{Y*},\mathbf{W}^{Z*},\mathbf{\Theta})$, where the optimal parameters $\mathbf{W}^{Y*}$ and $\mathbf{W}^{Z*}$ are obtained by minimizing the training losses as follows.
\begin{align}
    \small
    \mathbf{W}^{Y*} = \arg\min_{\mathbf{W}^Y} L_{train}^Y (\mathbf{W}^Y, \mathbf{\Theta}^*) \nonumber\\
    \mathbf{W}^{Z*} = \arg\min_{\mathbf{W}^Z} L_{train}^Z (\mathbf{W}^Z, \mathbf{\Theta}^*) \nonumber
\end{align}
The above goal forms a bi-level optimization problem~\cite{maclaurin2015gradient,pham2018efficient}, where $\mathbf{\Theta}$ is the upper-level variable and $\mathbf{W}^Y$ and $\mathbf{W}^Z$ are the lower-level variables: 
	\begin{equation}
	\begin{aligned}
	\label{equ:bilevel}
	\min_\mathbf{\Theta} \; &L_{val} \big(\mathbf{W}^{Y*}(\mathbf{\Theta}),\mathbf{W}^{Z*}(\mathbf{\Theta}),\mathbf{\Theta}\big)\\
	s.t. \; & \mathbf{W}^{Y*}(\mathbf{\Theta}) = \arg\min_{\mathbf{W}^Y} L_{train}^Y (\mathbf{W}^Y, \mathbf{\Theta}^*)\\
	 & \mathbf{W}^{Z*}(\mathbf{\Theta}) = \arg\min_{\mathbf{W}^Z} L_{train}^Z (\mathbf{W}^Z, \mathbf{\Theta}^*)
	\end{aligned}
	\end{equation}
\noindent Optimizing $\mathbf{\Theta}$ is time-consuming due to the expensive inner optimization of $\mathbf{W}^Y$ and $\mathbf{W}^Z$. Therefore, we leverage the approximation scheme as DARTS: 
\begin{equation}
\begin{aligned}
\label{equ:approximation}
\nabla_\mathbf{\Theta} \;L_{val} \big(&\mathbf{W}^{Y*}(\mathbf{\Theta}),\mathbf{W}^{Z*}(\mathbf{\Theta}),\mathbf{\Theta}\big)\\
\approx \nabla_\mathbf{\Theta} \;L_{val} \big(&\mathbf{W}^Y - \xi \nabla_{\mathbf{W}^Y}L_{train}^Y (\mathbf{W}^Y, \mathbf{\Theta}),\\
&\mathbf{W}^Z - \xi \nabla_{\mathbf{W}^Z}L_{train}^Z (\mathbf{W}^Z, \mathbf{\Theta}),\mathbf{\Theta}\big)\nonumber
\end{aligned}
\end{equation}
\noindent where $\xi$ is the learning rate for updating $\mathbf{W}^Y$ and $\mathbf{W}^Z$. The approximation scheme estimates $\mathbf{W}^{Y*}(\mathbf{\Theta})$ and $\mathbf{W}^{Z*}(\mathbf{\Theta})$ by updating $\mathbf{W}^Y$ and $\mathbf{W}^Z$ for a single training step, which avoids total optimization $\mathbf{W}^*(\mathbf{\Theta}) = \arg\min_\mathbf{W}$ $ L_{train} (\mathbf{W}, \mathbf{\Theta}^*)$ to the convergence. In our implementation, we apply first-order approximation with $\xi = 0$, which can even lead to more speed-up. Also, in our specific experiments, since the amount of validation data is limited, we build an augmented validation dataset $\mathcal{V}'=\mathcal{V} \cup \mathcal{T}$ combining the original validation set $\mathcal{V}$ with the training set $\mathcal{T}$ for optimizing $\Theta$. More details of the DARTS-based optimization algorithm are shown in Appendix \ref{app:opt}.



    



%% file: experiments.tex
\section{Experiment}
\label{sec:exp}
In this section, we conduct experiments to evaluate our proposed debiasing framework. Through the experiments, we try to answer two questions: 1) Does our framework effectively mitigate the implicit bias in various deep text classification models? and 2) Does our framework maintain the performance of the original models (without debasing) while reducing the bias?

\subsection{Baselines}
In our experiments, we compare our proposed debiasing framework with two baselines. Since there is no established method for mitigating implicit bias, we adopt two debiasing methods designed for traditional explicit bias and adapt them for implicit bias.

\textbf{Data Augmentation* (Data Aug)} \cite{dixon2018measuring}. We manually balance the training data of two demographic groups by adding sufficient negative examples for Group I and positive examples for Group II. As a result, the ratio of positive and negative training examples for both groups is 1:1.  As discussed in the introduction, obtaining additional labeled data from specific authors is very expensive. In this work, we seek to develop bias mitigation methodology without extra data. {\it Since Data Aug introduces more training data, it's not fair to directly compare it with other debiasing methods that only utilize original training data (including our method). We include Data Aug as a special baseline for reference.}

\textbf{Instance Weighting (Ins Weigh)} \cite{zhang2020demographics}. We re-weight each training instance with a numerical weight $\frac{P(Y)}{P(Y|Z)}$ based on the label distribution for each demographic group to mitigate explicit bias. In this method, a random forest classifier is built to estimate the conditional distribution $P(Y|Z)$ and the marginal distribution $P(Y)$ is manually calculated.


\begin{table*}[t]
\small
\centering
\caption{Fairness Performance Comparison. Note that Data Aug is a special baseline for reference.}
	\vspace{-0.05in}
\label{tab:fairness}
\begin{tabular}{c|c|ccc|ccc}
\hline
\hline
\multirow{2}{1cm}{\textbf{Task}} & \multirow{2}{1cm}{\textbf{Methods}} & \multicolumn{3}{c}{\textbf{\begin{tabular}[c]{@{}c@{}}CNN \end{tabular}}} & \multicolumn{3}{|c}{\textbf{\begin{tabular}[c]{@{}c@{}}RNN \end{tabular}}}\\ \cline{3-5}  \cline{6-8}
 & & \textbf{FPED (\%)} & \textbf{FNED (\%)} & \textbf{DPD (\%)} & \textbf{FPED (\%)} & \textbf{FNED (\%)} & \textbf{DPD (\%)}\\ \hline
 & \textbf{Base Model} & 23.59 & 41.45 & 32.52 & 26.86 & 42.36 & 34.61 \\ 
\textbf{Sentiment} & \textbf{Data Aug*} & 21.00* & 3.88* & 12.44* & 19.84* & 0.59* & 10.22*  \\
\textbf{Race} & \textbf{Ins Weigh} & 25.47 & 41.43 & 33.45 & 26.86 & 42.36 & 34.61 \\ 
\textbf{(DIAL)} & \textbf{Debiased-TC} & 6.08 & 4.63 & 0.73 & 6.67 & 5.68 & 0.50 \\ 
\hline
\hline
 & \textbf{Base Model} & 32.73 & 17.58 & 25.16 & 30.44 & 17.55 & 24.00 \\ 
\textbf{Mention} & \textbf{Data Aug*} & 1.31* & 7.31* & 3.00* & 0.77* & 7.91* & 4.34*  \\ 
\textbf{Race} & \textbf{Ins Weigh} & 24.66 & 19.46 & 22.06 & 28.83 & 17.26 & 23.05  \\ 
\textbf{(DIAL)} & \textbf{Debiased-TC} & 3.61 & 2.40 & 0.61 & 4.97 & 1.07 & 1.95 \\ 
\hline
\hline
 & \textbf{Base Model} & 11.60 & 9.95 & 10.78 & 10.62 & 8.33 & 9.47 \\ 
\textbf{Mention} & \textbf{Data Aug*} & 0.84* & 0.19* & 0.32* & 2.42* & 0.72* & 1.57* \\ 
\textbf{Gender} & \textbf{Ins Weigh} & 12.73 & 10.22 & 11.47 & 11.20 & 9.35 & 10.28 \\ 
\textbf{(PAN16)} & \textbf{Debiased-TC} & 3.95 & 3.04 & 3.49 & 5.41 & 3.73 & 4.57 \\ 
\hline
\hline
 & \textbf{Base Model} & 15.03 & 9.96 & 12.49 & 13.07 & 7.34 & 10.20 \\ 
\textbf{Mention} & \textbf{Data Aug*} & 3.71* & 1.59* & 1.06* & 0.17* & 2.69* & 1.26* \\ 
\textbf{Age} & \textbf{Ins Weigh} & 16.53 & 8.71 & 12.62 & 13.24 & 7.94 & 10.59 \\ 
\textbf{(PAN16)} & \textbf{Debiased-TC} & 7.29 & 2.91 & 5.10 & 7.64 & 2.69 & 5.16 \\ 
\hline
\hline
 & \textbf{Base Model} & 78.56 & 37.22 & 57.89 & 81.51 & 28.50 & 55.01 \\ 
\textbf{Hate Speech} & \textbf{Data Aug*} & 88.81* & 26.15* & 57.48* & 83.51* & 22.73* & 53.12*  \\ 
\textbf{Race} & \textbf{Ins Weigh} & 87.51 & 31.92 & 59.72 & 84.45 & 27.44 & 55.95 \\ 
\textbf{(MTC)} & \textbf{Debiased-TC} & 75.97 & 17.08 & 46.53 & 74.56 & 18.85 & 46.70 \\ 
\hline
\hline

\end{tabular}
\end{table*}

\subsection{Experimental Settings}
\label{sec:settings}
We conduct our experiments for implicit bias mitigation on two representative base models: \textbf{CNN} \cite{kim2014convolutional} and \textbf{RNN} \cite{chung2014empirical}. We use the same datasets with manually designed proportions, as described in Section \ref{sec:prelim_results}. The details of the base models, as well as the implementation details for the replication of the experiments, can be found in Appendix \ref{app:imp}.

\subsection{Performance Comparison}

We train the base models with our proposed debiasing framework as well as the baseline debiasing methods. We report the performance on the test set in terms of fairness and classification performance. 

\noindent\textbf{Fairness Evaluation.} Table \ref{tab:fairness} shows the results for fairness evaluation metrics: false positive equality difference (FPED), false negative equality difference (FNED), and DPD. FPED/FNED indicates the absolute value of the difference between the false positive/negative rates of the two groups. We make the following observations. First, the base models attain high FPED, FNED, and DPD, which indicates the existence of significant implicit bias towards the authors of the texts. Ins Weigh seems ineffective in mitigating implicit bias since it only achieved comparable fairness scores with the base models. Note that not every example that belongs to a certain group necessarily results in bias towards that group. Thus, assigning a uniform weight for all examples with the same label $Y$ and demographic attribute $Z$ is not a proper way to reduce implicit bias. Third, both Data Aug and Debiased-TC can mitigate the implicit bias by achieving lower equality and demographic parity differences. However, compared to Data Aug, Debiased-TC has two advantages.  First, Data Aug needs to add more training data while Debiased-TC does not. Debiased-TC can locate the main source of implicit bias by analyzing how it forms in a deep text classification model. Due to the proposed corrector model, it can make a classification model focus on the relevant features for predictions and discard the features that may lead to implicit bias. Second, Debiased-TC is more stable than Data Aug. For the sentiment classification task with race as the demographic attribute, the CNN and RNN classifiers trained on augmented data still result in high FPED and DPD scores. This suggests that balancing the training data cannot always mitigate implicit bias. In fact, only training examples with demographic language features can contribute to the implicit bias. Since some texts in the training set do not contain any language features belonging to a demographic group, they do not help balance the data.



\begin{table*}[t]
\small
\centering
\caption{Text Classification Performance Comparison (\%). Note that Data Aug is a  special baseline for reference.}
	\vspace{-0.1in}
\label{tab:performance}
\begin{tabular}{ccccccccccc}
\hline
\hline
\multirow{3}{1cm}{\textbf{Methods}} & \multicolumn{2}{c}{\textbf{\begin{tabular}[c]{@{}c@{}}Sentiment/Race \end{tabular}}} & \multicolumn{2}{c}{\textbf{\begin{tabular}[c]{@{}c@{}}Mention/Race \end{tabular}}} & \multicolumn{2}{c}{\textbf{\begin{tabular}[c]{@{}c@{}}Mention/Gender \end{tabular}}} & \multicolumn{2}{c}{\textbf{\begin{tabular}[c]{@{}c@{}}Mention/Age \end{tabular}}} & \multicolumn{2}{c}{\textbf{\begin{tabular}[c]{@{}c@{}}Hate Speech/Race \end{tabular}}}\\
 & \multicolumn{2}{c}{\textbf{\begin{tabular}[c]{@{}c@{}}(DIAL) \end{tabular}}} & \multicolumn{2}{c}{\textbf{\begin{tabular}[c]{@{}c@{}}(DIAL) \end{tabular}}} & \multicolumn{2}{c}{\textbf{\begin{tabular}[c]{@{}c@{}}(PAN16) \end{tabular}}} & \multicolumn{2}{c}{\textbf{\begin{tabular}[c]{@{}c@{}}(PAN16) \end{tabular}}} & \multicolumn{2}{c}{\textbf{\begin{tabular}[c]{@{}c@{}}(MTC) \end{tabular}}}\\ \cmidrule(r){2-3}  \cmidrule(r){4-5} \cmidrule(r){6-7} \cmidrule(r){8-9} \cmidrule(r){10-11}
 & \textbf{Acc.} & \textbf{F1} & \textbf{Acc.} & \textbf{F1} & \textbf{Acc.} & \textbf{F1} & \textbf{Acc.} & \textbf{F1} & \textbf{Acc.} & \textbf{F1}\\ \hline
 & \multicolumn{10}{c}{\textbf{\begin{tabular}[c]{@{}c@{}} CNN \end{tabular}}}\\
\hline
\textbf{Base Model} & 61.40 & 60.03 & 70.77 & 71.65 & 81.93 & 81.94 & 80.57 & 80.17 & 64.10 & 65.86\\ 
 \textbf{Data Aug*} & 67.58* & 71.53* & 76.42* & 76.03* & 84.11* & 84.31* & 84.08* & 84.36* & 66.96* & 71.10*\\ 
\textbf{Ins Weigh} & 61.06 & 60.36 & 71.62 & 69.66 & 81.86 & 81.85 & 80.70 & 81.05 & 65.25 & 68.73\\ 
\textbf{Debiased-TC} & 63.60 & 66.58 & 73.15 & 71.84 & 81.67 & 82.01 & 80.41 & 79.68 & 69.14 & 72.69\\ 
\hline
 & \multicolumn{10}{c}{\textbf{\begin{tabular}[c]{@{}c@{}} RNN \end{tabular}}}\\ \hline
\textbf{Base Model} & 61.23 & 61.53 & 72.97 & 73.68 & 83.46 & 83.40 & 82.78 & 82.43 & 66.31 & 69.57\\ 
 \textbf{Data Aug*} & 67.82* & 69.35* & 78.42* & 77.26* & 86.25* & 86.05* & 86.12* & 85.68* & 68.55* & 72.37*\\ 
\textbf{Ins Weigh} & 61.23 & 61.53 & 73.37 & 73.79 & 83.46 & 83.32 & 82.80 & 82.58 & 67.26 & 70.94\\ 
\textbf{Debiased-TC} & 63.68 & 66.70 & 74.05 & 73.41 & 81.81 & 81.51 & 80.21 & 79.17 & 66.76 & 70.76\\ 
\hline
\hline
\end{tabular}
	\vspace{-0.2in}
\end{table*}


\noindent\textbf{Text Classification Performance Evaluation.} The prediction performance of the text classification models trained under various debiasing methods is shown in Table \ref{tab:performance}, where we report the accuracy and F1 scores. First, it is not surprising to see that Data Aug achieves the best performances, since the data augmentation technique introduces more training data. It's not fair to directly compare it with other debiasing methods that only utilize original training data.  Second, in most cases, our method achieves comparable or even better performance than the original base models. As we verified before, the implicit bias of a text classification model is caused by the fact that it learns a wrong correlation between labels and demographic language features. Debiased-TC corrects the model's selection of language features for predictions and thereby improves its performance on the classification task.


In conclusion, our proposed debiasing framework significantly mitigates the implicit bias, while maintaining or even slightly improve the classification performance.

%% file: relatedwork.tex
\section{Related Work}
\label{sec:related}

\textbf{Fairness in NLP.} Recent research has demonstrated that word embeddings exhibit human biases for text data. For example, in word embeddings trained on large-scale real-world text data, the word ``man'' is mapped to ``programmer'' while ``woman'' is mapped to ``homemaker''~\cite{NIPS2016_6228}.
Some works extend the research of biases in word embeddings to that of sentence embeddings. The work~\cite{DBLP:journals/corr/abs-1903-10561} examines popular sentence encoding models from CBoW, GPT, ELMo to BERT, and shows that those models inherit human's prejudices from the training data. For the task of coreference resolution, a benchmark named WinoBias is proposed \cite{DBLP:journals/corr/abs-1804-06876} to measure the gender biases with a debiasing method based on data augmentation.
\citet{DBLP:journals/corr/abs-1809-02208} reveal that Google's machine translation system shows gender biases in various languages. Existing debiasing methods for word embeddings are adopted to mitigate the biases in machine translation systems \cite{DBLP:journals/corr/abs-1904-03035}.
In the task of dialogue generation,  it is first studied by \cite{liu2020does} on the biases learned by dialogue agents from human conversation data. It is shown that significant gender and race biases exist in popular dialogue models. As a countermeasure, \citet{liu2020mitigating} propose to mitigate gender bias in neural dialogue models with adversarial learning. 

\textbf{Fairness in Text Classification.} For the text classification problem, \citet{dixon2018measuring} demonstrate that the source of unintended bias in models is the imbalance of training data, and they provide a debiasing method, which introduces new data to balance the training data. In~\cite{park2018reducing}, gender biases are measured on abusive language detection models, and the effect of different pre-trained word embeddings and model architectures are analyzed.
By considering the various ways that a classifier’s score distribution can vary across designated groups, a suite of threshold-agnostic metrics is introduced in \cite{borkan2019nuanced}, which provides a nuanced view of this unintended bias. Furthermore, the work~\cite{zhang2020demographics} proposes to debias text classification models using instance weighting, i.e., different weights are assigned to the training samples involving different demographic groups. The works discussed above focus on explicit bias, where the demographic attributes are explicitly expressed in the text. However, works studying implicit bias are rather limited. \citet{huang2020multilingual} introduce the first multilingual hate speech dataset with inferred author demographic attributes. Through experiments on this dataset, they show that popular text classifiers can learn the bias towards the demographic attribute of the author. But this work doesn't discuss how the bias is produced, and no debiasing method is provided.

%% file: conclusion.tex
\section{Conclusion}
\label{sec:con}

In this paper, we demonstrate that a text classifier with implicit bias makes predictions based on language features correlated with demographic groups of authors, and propose a novel learning framework Debiased-TC to mitigate such implicit bias.
%
The experimental results show that Debiased-TC significantly mitigates implicit bias, and maintains or even improves the text classification performance of the original models. In the future, we will investigate implicit bias in other NLP applications. 

%% file: appendix.tex
\section{Data Statistics}
\label{app:sta}

The statistics of the datasets DIAL, PAN16, and MTC are shown in Table~\ref{tab:stat}. In the table, the ``task'' section shows the text classification tasks included in a dataset. ``Sentiment'' is short for sentiment analysis. ``Mention'' is short for mention detection. ``Hate Speech'' is short for hate speech detection. ``Demog'' indicates the demographic attribute of the tweet authors collected in a dataset. The ``Size'' section shows the total number of instances in a dataset. Each instance is a tweet text. The ``Avg.Len'' section shows the average number of words in one instance in a dataset. The datasets DIAL and PAN16 can be download from the link \footnote{\url{https://github.com/yanaiela/demog-text-removal}}. The dataset MTC can be downloaded from the link \footnote{\url{https://github.com/xiaoleihuang/Multilingual_Fairness_LREC}}.

\begin{table}[h]
    \small
	\centering
	\caption{Statistics of the datasets.}
	\begin{tabular}{c|c|c|c|c}
	    \hline \hline
		Dataset & Task & Demog & Size   & Avg.Len\\ \hline
		\multirow{2}{1cm}{DIAL} & Sentiment & Race  & 317,151  & 11.20 \\
        & Mention & Race  & 400,000 & 10.56 \\
        \hline
		\multirow{2}{1cm}{PAN16} &  Mention & Gender & 175,871 & 14.64  \\
     & Mention & Age  & 175,471   & 14.55  \\
     \hline
	 MTC & Hate Speech & Race & 47,627 & 19.60  \\
		\hline \hline
	\end{tabular}
	\label{tab:stat}
\end{table}

\section{Optimization of the Explainer}
\label{app:exp}
We train the explainer $\mathcal{E}$ by maximizing the mutual information between the response variable $Y$ and the selected features $X_S$. The optimization problem can be formulated as:
\begin{align}
\label{eq:optim}
  \max_{\mathcal{E}} \quad & I(X_S;Y)\\
  \text{s.t.} \quad & S \sim P_{\mathcal{E}}(S|X) \nonumber
\end{align}
\noindent where 
\begin{align}
I(X_S,Y) &= \mathbb{E} \Big[ \log \frac{P(X_S,Y)}{P(X_S) P(Y)} \Big] \nonumber\\
& = \mathbb{E} \Big[ \log \frac{P_{\mathcal{M}}(Y | X_S)}{P(Y)} \Big]  \nonumber\\
& \propto \mathbb{E} \Big[\log P_{\mathcal{M}}(Y|X_S)\Big ]  \nonumber\\
& = \mathbb{E}_X \mathbb{E}_{S|X} \mathbb{E}_{Y|X_S} \Big [\log P_{\mathcal{M}}(Y|X_S) \Big] \nonumber
\end{align}

Solving the optimization problem in Eq.~(\ref{eq:optim}) is equivalent to finding an explainer $\mathcal{E}$ satisfying the following:
\begin{align}
\label{eq:final_optim}
  \max_{\mathcal{E}} P_{\mathcal{M}}(Y|X_S) \quad \text{s.t.} \qquad S \sim P_{\mathcal{E}}(S|X). \nonumber
\end{align}

Hence, we train the explainer $\mathcal{E}$ by optimizing $P_{\mathcal{M}}(Y|X_S)$ with the parameters of the classification model $\mathcal{M}$ fixed. In our implementation, we adopt the cross-entropy loss for training, as we do when we train the classification model $\mathcal{M}$.

\section{An Optimization Method for Debiased-TC}
\label{app:opt}

We present our DARTS-based optimization algorithm in Algorithm \ref{alg:opt}. In each iteration, we first update the corrector's parameters based on the augmented validation set $\mathcal{V}'$ (lines 2-3). Then, we collect a new mini-batch of training data (line 4). We generate the saliency scores $S=(s_1, s_2, \dots, s_n)$ for the training examples via the corrector with its current parameters (line 5). Next, we make predictions via the classifiers with their current parameters and $X_{S}$ (line 6).  Eventually, we update the parameters of the classifiers (line 7).

\begin{algorithm}[h]
    \small
	\caption{\label{alg:opt} The DARTS-based optimization method for Debiased-TC.}
	\raggedright
	{\bf Input}: Training data $\mathcal{T}=\{X_i, Y_i, Z_i\}_{i=1}^{|\mathcal{T}|}$ and Validation data $\mathcal{V}=\{X_i, Y_i, Z_i\}_{i=1}^{|\mathcal{V}|}$\\
	{\bf Output}: classifier parameters $\mathbf{W}^{Y*}$ and $\mathbf{W}^{Z*}$; and corrector parameters $\mathbf{\Theta}^*$\\
	Initialize $\mathbf{W}^{Y}$, $\mathbf{W}^{Z}$ and $\mathbf{\Theta}$\\
	\begin{algorithmic} [1]
		\WHILE{not converged}
		\STATE Sample a mini-batch of validation data from $\mathcal{V}'=\mathcal{V} \cup \mathcal{T}$
		\STATE Update $\mathbf{\Theta}$ by descending $\nabla_\mathbf{\Theta} \;L_{val} \big(\mathbf{W}^Y - \xi \nabla_{\mathbf{W}^Y}L_{train}^Y (\mathbf{W}^Y, \mathbf{\Theta}),$\\
        $\qquad \qquad \mathbf{W}^Z - \xi \nabla_{\mathbf{W}^Z}L_{train}^Z (\mathbf{W}^Z, \mathbf{\Theta}),\mathbf{\Theta}\big)$ \\
        ($\xi = 0$ for first-order approximation)\\
		\STATE Collect a mini-batch of training data from $\mathcal{T}$
		\STATE Generate $S$ via the corrector with current parameters $\mathbf{\Theta}$
		\STATE Generate predictions via the classifiers with current parameters $\mathbf{W}^Y$, $\mathbf{W}^Z$ and $X_{S}$
		\STATE Update $\mathbf{W}^Y$ and $\mathbf{W}^Z$ by descending $\nabla_{\mathbf{W}^Y}L_{train}^Y (\mathbf{W}^Y, \mathbf{\Theta})$ and $\nabla_{\mathbf{W}^Z}L_{train}^Z (\mathbf{W}^Z, \mathbf{\Theta})$
		\ENDWHILE
	\end{algorithmic}
\end{algorithm}

\section{Implementation Details}
\label{app:imp}
\subsection{Details of Base Models}
In the base model CNN, we use 100 filters with three different kernel sizes (3, 4, and 5) in the convolution layer, where we use a Rectified Linear Unit (ReLU) as the non-linear activation function. Each obtained feature map is processed by a max-pooling layer. Then, the features are concatenated and fed into a linear prediction layer to get the final predictions. A dropout with a rate of 0.3 is applied before the linear prediction layer.

For the base model RNN, we use a one-layer unidirectional RNN with Gated Recurrent Units (GRU). The hidden size is set to 300. The last hidden state of the RNN is fed into a linear prediction layer to get the final predictions. We apply a dropout with a rate of 0.2 before the linear prediction layer.

\subsection{Details of Experimental Settings}
For the text classifiers, we use randomly initialized word embeddings with a size of 300. All the models are trained by an Adam optimizer \cite{kingma2014adam} with an initial learning rate of 0.001. We apply gradient clipping with a clip-value of 0.25 to prevent the exploding gradient problem. The batch size is set to 64. For the base model and the baseline methods, when the prediction accuracy of the validation data doesn't improve for 5 consecutive epochs, the training is terminated, and we pick the model with the best performance on the validation set. Our model utilizes the validation data for training. To avoid it overfitting the validation data, we don't select the model based on its performance on the validation set. Instead, we train the model for a fixed number of epochs (5 epochs, the same for all the three datasets) and evaluate the obtained model. 





